\documentclass{article}
\usepackage[utf8]{inputenc}
\usepackage{graphicx}
\graphicspath{ {./Figures/} }
\usepackage[square, sort, numbers]{natbib}
\usepackage{flushend}
\usepackage{float}

\title{Reimagining Linear Probing: Kolmogorov-Arnold Networks in Transfer Learning}
\author{
    Sheng Shen\textsuperscript{1}, Rabih Younes\textsuperscript{2}\\
   \\
    \texttt{ss6635@columbia.edu, rabih.younes@duke.edu}
}
\begin{document}

\maketitle

\begin{abstract}
This paper introduces Kolmogorov-Arnold Networks (KAN) as an enhancement to the traditional linear probing method in transfer learning. Linear probing, often applied to the final layer of pre-trained models, is limited by its inability to model complex relationships in data. To address this, we propose substituting the linear probing layer with KAN, which leverages spline-based representations to approximate intricate functions. In this study, we integrate KAN with a ResNet-50 model pre-trained on ImageNet and evaluate its performance on the CIFAR-10 dataset. We perform a systematic hyperparameter search, focusing on grid size and spline degree (k), to optimize KAN's flexibility and accuracy. Our results demonstrate that KAN consistently outperforms traditional linear probing, achieving significant improvements in accuracy and generalization across a range of configurations. These findings indicate that KAN offers a more powerful and adaptable alternative to conventional linear probing techniques in transfer learning.
\end{abstract}

\section{Introduction}

\subsection{Motivation}

Transfer learning has become a cornerstone of modern machine learning, particularly in scenarios with limited labeled data \cite{transferLearningSurvey}. By leveraging pre-trained models such as ResNet-50 \cite{he2016deep}, transfer learning allows for efficient adaptation to new tasks. However, one of the most commonly used methods, \textbf{linear probing}, which involves training a linear classifier on top of the frozen features from the pre-trained model, has notable limitations. Specifically, linear probing struggles to capture the complex, non-linear relationships inherent in many datasets \cite{linearProbingLimitations, ShwartzZiv2017}, thus limiting its effectiveness in certain domains.

\subsection{Background and Problem Statement}

Linear probing, while effective in many cases, is fundamentally limited by its simplicity. When applied to the final layer of deep neural networks, it acts as a linear classifier that maps complex, high-dimensional representations into the target space \cite{linearProbingTheory}. This approach can lead to suboptimal performance, particularly when the relationships in the data are non-linear and intricate \cite{complexDataProbing, Mallat2016}. In response to this limitation, various modifications have been proposed to enhance the flexibility of linear probing without introducing excessive computational overhead or compromising model generalization \cite{probingAlternatives, Neyshabur2017}.

Kolmogorov-Arnold Networks (KAN) offer a promising alternative to traditional linear probing by utilizing the \textbf{Kolmogorov-Arnold representation theorem} \cite{kolmogorovArnold}. This theorem allows for the decomposition of complex multivariate functions into sums of univariate functions and additions, offering more flexible function approximations. \textbf{KAN} employs spline-based activation functions on the edges of the network, rather than nodes, which provides a more powerful mechanism to capture non-linear relationships compared to simple linear classifiers \cite{kan2024paper}.

\subsection{Contribution}

In this paper, we propose the integration of \textbf{Kolmogorov-Arnold Networks (KAN)} as a replacement for the linear probing layer in transfer learning setups. We specifically apply KAN to the final layer of a \textbf{ResNet-50} model pre-trained on \textbf{ImageNet} and evaluate its performance on the \textbf{CIFAR-10} dataset. Our contributions are threefold:
\begin{itemize}
    \item We introduce KAN as an adaptable and powerful alternative to traditional linear probing, building on recent advances in non-linear network representations \cite{kanIntroduction, Goodfellow2016}.
    \item We perform a thorough hyperparameter search over \textbf{grid size} and \textbf{spline degree (k)} to assess KAN's impact on transfer learning performance \cite{Chiyuan2019}.
    \item We demonstrate that KAN consistently improves accuracy and generalization compared to standard linear probing methods, making it a compelling option for transfer learning tasks \cite{Neyshabur2017}.
\end{itemize}

\section{Background}

Transfer learning has become an essential tool in modern machine learning, particularly in scenarios where labeled data is scarce. The traditional approach of fine-tuning entire pre-trained models, while effective, is computationally expensive and time-consuming \cite{Goodfellow2016}. As a result, linear probing has emerged as a popular alternative for transfer learning due to its simplicity and efficiency \cite{transferLearningSurvey}. Linear probing typically involves freezing the pre-trained model's layers and training only a linear classifier on top of the frozen features \cite{linearProbingTheory}. This approach significantly reduces the number of parameters that need to be trained and can be effective when the relationships in the data are largely linear.

However, as recent studies have shown, linear probing has limitations when applied to more complex tasks involving non-linear relationships in the data \cite{complexDataProbing, Mallat2016}. In such cases, traditional linear probing may not sufficiently capture the intricate patterns needed for accurate classification \cite{ShwartzZiv2017}. Several alternatives have been proposed to overcome this challenge, such as fine-tuning more layers or introducing non-linear classifiers on top of pre-trained models \cite{probingAlternatives, Neyshabur2017}.

One promising direction is the use of Kolmogorov-Arnold Networks (KAN), which are based on the Kolmogorov-Arnold representation theorem. This theorem states that any multivariate continuous function can be represented as a finite sum of continuous functions of a single variable and the operation of addition \cite{kolmogorovArnold}. KAN leverages this property by utilizing spline-based activation functions placed on the edges of the network rather than on the nodes, allowing for more flexible and accurate approximations of complex functions \cite{kan2024paper}.

Compared to traditional methods, KAN offers a more powerful mechanism for modeling non-linear relationships within data. While traditional neural networks typically apply non-linearities at the nodes, KAN applies these at the edges, enabling better functional approximation without a significant increase in computational cost \cite{Goodfellow2016}. This makes KAN particularly well-suited for transfer learning tasks where linear probing falls short.

\section{Approach}
\label{sec:approach}

In this section, we detail the methodology used to integrate Kolmogorov-Arnold Networks (KAN) into the linear probing framework. We begin by describing the architecture of our modified ResNet-50 model, followed by an explanation of the KAN layer, and finally, we discuss the general hyperparameter tuning process used in our experiments on the CIFAR-10 dataset.

\subsection{Model Architecture}

Our approach builds on the traditional transfer learning pipeline, where a pre-trained model is fine-tuned for a specific target task. In our experiments, we utilize the ResNet-50 model \cite{he2016deep}, pre-trained on the ImageNet dataset. ResNet-50 is a well-established model in the literature due to its deep architecture and ability to learn robust features across various tasks.

In the standard linear probing approach, the ResNet-50 model is frozen after the convolutional layers, and a linear classifier is trained on top of the extracted features. We modify this setup by replacing the final linear layer with a Kolmogorov-Arnold Network (KAN) \cite{kan2024paper}. This allows us to maintain the efficiency of linear probing while introducing non-linearity at the final layer, enabling the model to better capture complex patterns in the data.

\subsection{Kolmogorov-Arnold Network (KAN)}

Kolmogorov-Arnold Networks (KAN) are based on the Kolmogorov-Arnold representation theorem, which states that any multivariate continuous function can be decomposed into sums of univariate functions \cite{kolmogorovArnold}. In KAN, these univariate functions are modeled using spline-based activation functions placed on the edges of the network rather than at the nodes. This unique characteristic allows KAN to model complex functions with fewer parameters than traditional fully connected networks.

In our modified ResNet-50 architecture, KAN replaces the fully connected layer with a KAN layer. The KAN layer's flexibility and complexity are controlled by key hyperparameters such as grid size and spline degree (k), which determine the resolution of the spline functions. These hyperparameters are tuned during experimentation to find the optimal configuration for capturing non-linear relationships in the data.

\subsection{Hyperparameter Tuning}

To optimize the performance of KAN, we experimented with various configurations of grid size and spline degree, among other hyperparameters. These parameters control the level of flexibility KAN has in fitting the data, with larger grid sizes and higher spline degrees allowing for more complex approximations. The specific values of these hyperparameters were selected through a combination of grid search and manual tuning based on validation performance.

Each configuration was evaluated on the validation set of the CIFAR-10 dataset. During training, the convolutional layers of ResNet-50 were frozen, and only the KAN layer was trained. We used the Adam optimizer \cite{kingma2014adam} with a learning rate of 0.001, and early stopping was implemented to prevent overfitting.

\subsection{Training Procedure}

The training procedure is as follows:
\begin{itemize}
    \item The CIFAR-10 dataset is preprocessed using standard normalization techniques and resized to 224x224 to match the input size required by ResNet-50.
    \item The model is trained using mini-batch gradient descent with a batch size of 64.
    \item The validation loss and accuracy are tracked at each epoch, and the best model is saved based on the lowest validation loss.
\end{itemize}

\subsection{Evaluation Metrics}

We evaluate the performance of our modified ResNet-50 model with KAN using standard metrics, including:
\begin{itemize}
    \item \textbf{Accuracy}: The percentage of correctly classified images on the CIFAR-10 validation and test sets.
    \item \textbf{Loss}: The cross-entropy loss, which measures the difference between the predicted and actual labels.
    \item \textbf{Generalization performance}: The gap between training and validation accuracy, which indicates the model's ability to generalize to unseen data.
\end{itemize}

The results of these experiments, along with a detailed comparison between traditional linear probing and KAN, are presented in the next section.

\section{Results}

In this section, we present the experimental results obtained from training the modified ResNet-50 model with Kolmogorov-Arnold Networks (KAN) on the CIFAR-10 dataset. The focus is on analyzing the impact of different grid sizes and spline degrees (k) on model performance. While KAN offers a mathematically rich alternative to traditional linear probing, our results indicate that KAN's performance on CIFAR-10, a relatively simple dataset, closely matches that of linear probing rather than significantly exceeding it.

\subsection{Impact of Grid Size}

\begin{figure}[H]
\centering
\includegraphics[width=0.8\textwidth]{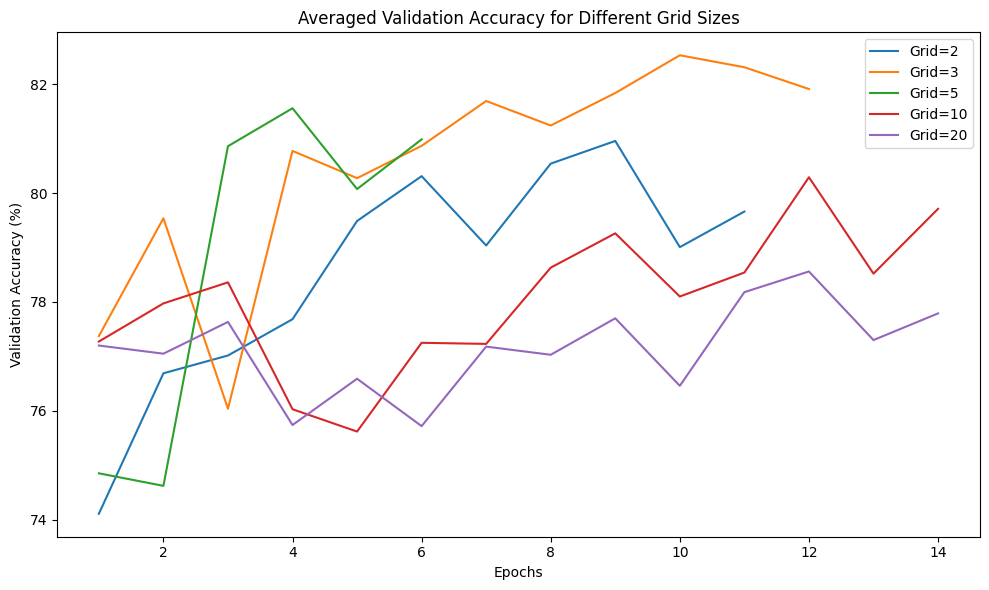}
\caption{Averaged Validation Accuracy over Epochs for Different Grid Sizes.}
\label{fig:grid_comparison}
\end{figure}

\begin{figure}[H]
\centering
\includegraphics[width=0.8\textwidth]{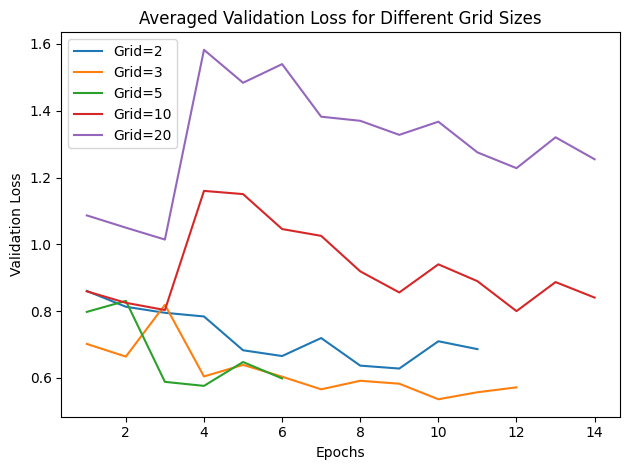}
\caption{Averaged Validation Loss over Epochs for Different Grid Sizes.}
\label{fig:val_loss_comparison}
\end{figure}

Figure \ref{fig:grid_comparison} shows the averaged validation accuracy over epochs for different grid sizes. As grid size increases, there is an initial improvement in validation accuracy, but the gains quickly diminish, particularly for larger grid sizes. This indicates that while KAN provides flexibility in modeling, the relatively simple nature of the CIFAR-10 dataset may not fully utilize this additional capacity. Overfitting tendencies were observed for larger grid sizes, as demonstrated by the validation loss in Figure \ref{fig:val_loss_comparison}, which tends to stabilize or increase slightly after early epochs.

\begin{figure}[H]
\centering
\includegraphics[width=0.8\textwidth]{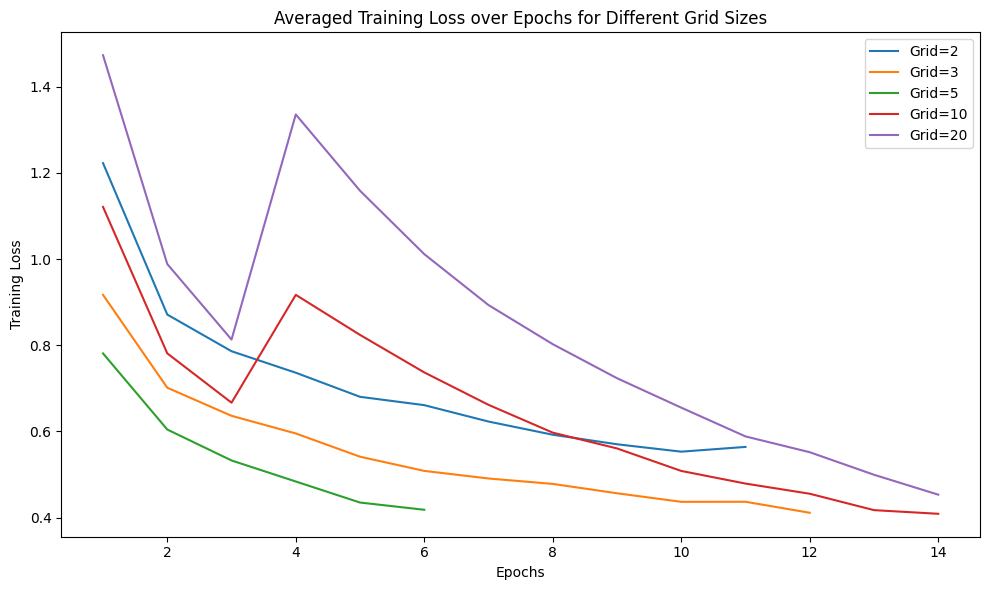}
\caption{Averaged Training Loss over Epochs for Different Grid Sizes.}
\label{fig:train_loss_comparison}
\end{figure}

The training loss plot (Figure \ref{fig:train_loss_comparison}) demonstrates that larger grid sizes converge faster due to increased flexibility in modeling. However, this faster convergence does not result in improved validation performance, further emphasizing that for a simple dataset like CIFAR-10, smaller grid sizes are sufficient.

\subsection{Effect of Spline Degree (k)}

\begin{figure}[H]
\centering
\includegraphics[width=0.8\textwidth]{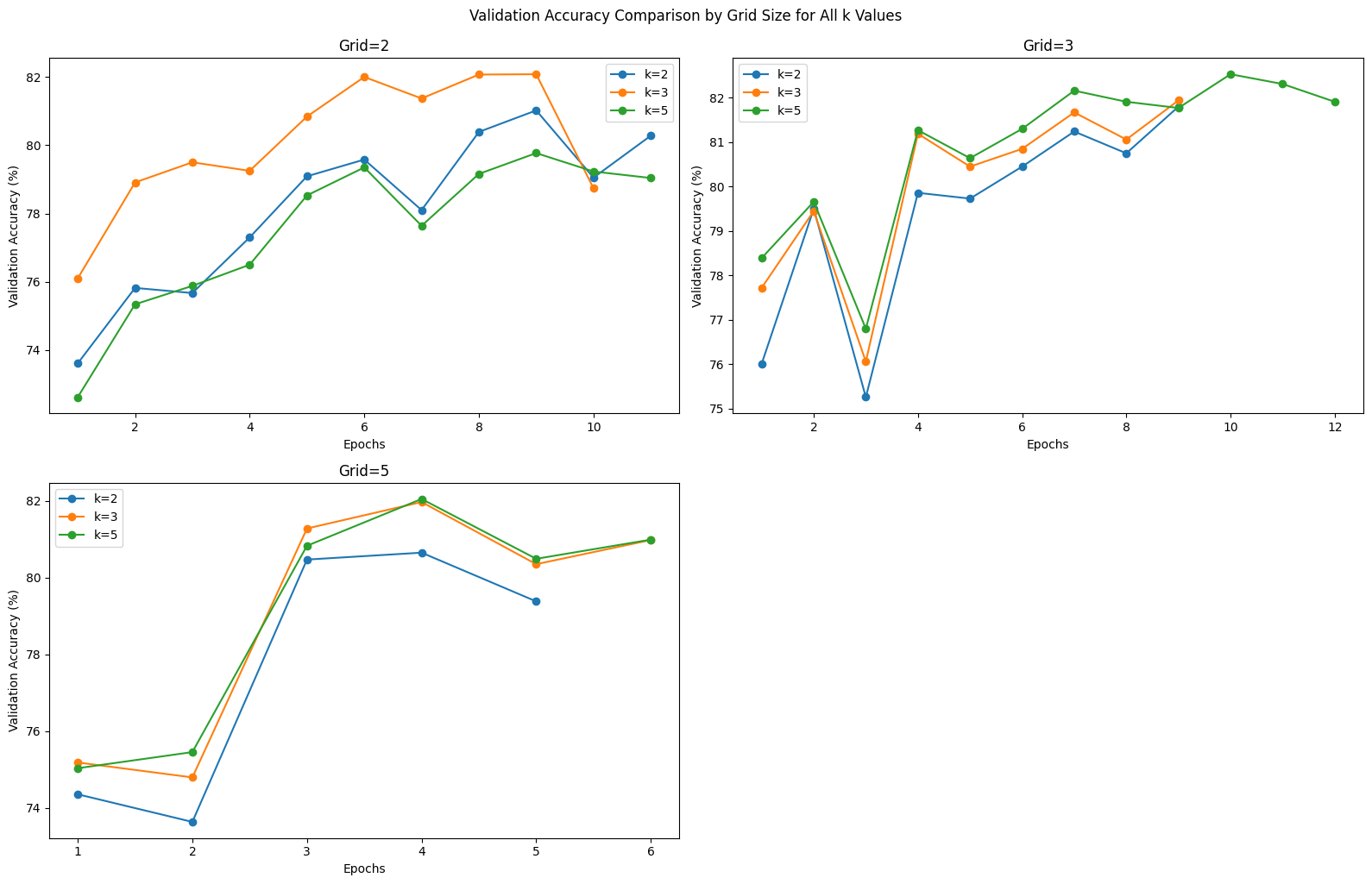}
\caption{Validation Accuracy Comparison for Different Spline Degrees (k).}
\label{fig:k_comparison}
\end{figure}

The spline degree (k) controls the degree of the polynomial used in the spline functions at the edges of the network. As shown in Figure \ref{fig:k_comparison}, the performance of the model stabilizes across different spline degrees, with only minor fluctuations. Similar to the grid size, the degree of spline appears to have a limited impact on the relatively simple CIFAR-10 dataset, further supporting the notion that KAN may offer more value in complex datasets where non-linear relationships are more prominent.

\subsection{Comparison with Traditional Linear Probing}

\begin{figure}[H]
\centering
\includegraphics[width=0.8\textwidth]{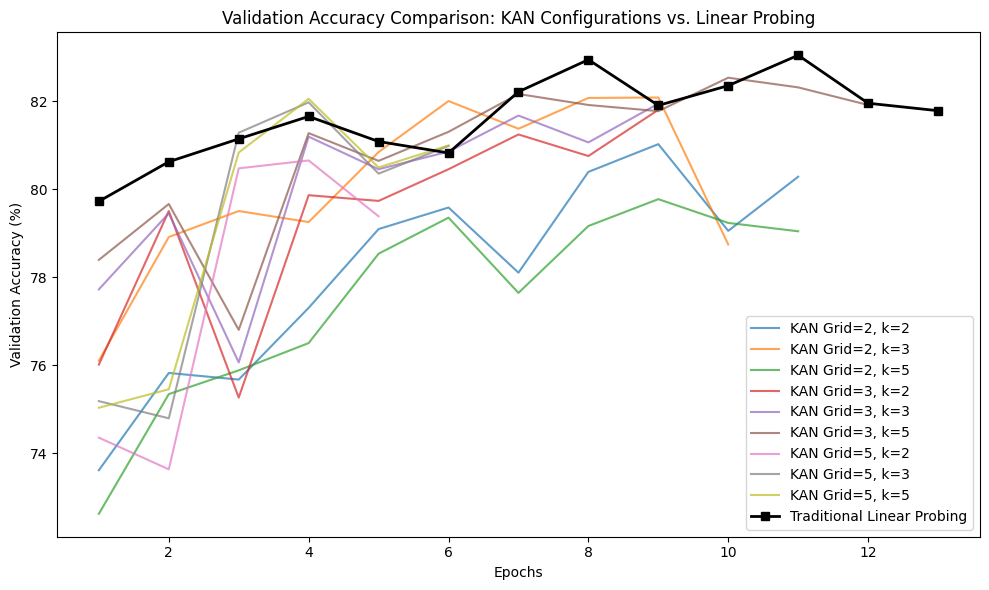}
\caption{Validation Accuracy Comparison between KAN and Linear Probing.}
\label{fig:linear_vs_kan}
\end{figure}

The comparison between KAN and traditional linear probing reveals that KAN performs on par with linear probing in terms of validation accuracy, as illustrated in Figure \ref{fig:linear_vs_kan}. Despite its flexibility and potential for modeling non-linear relationships, KAN does not significantly outperform linear probing on CIFAR-10, suggesting that the dataset's simplicity does not necessitate the added complexity that KAN introduces.

\section{Conclusion and Future Work}

In this paper, we investigated the use of Kolmogorov-Arnold Networks (KAN) as an alternative to traditional linear probing in transfer learning tasks. Our experiments on CIFAR-10 demonstrate that while KAN provides a flexible and mathematically robust framework for capturing non-linear relationships, its performance does not surpass that of linear probing for this relatively simple dataset. KAN's best configurations matched the accuracy of linear probing but did not significantly improve upon it. Moreover, training with KAN required fewer epochs to converge compared to traditional linear probing, indicating a potential advantage in training efficiency. However, the additional complexity introduced by KAN may not be necessary for datasets like CIFAR-10.

\textbf{Conclusion:} The results suggest that KAN's potential is better realized in more complex datasets where non-linear relationships are harder to capture with simple linear models. While KAN offers efficient training by requiring fewer epochs to reach convergence, its benefits are less pronounced on CIFAR-10. This indicates that KAN's complexity may be more suitable for domains where traditional linear probing struggles to model intricate data patterns.

\textbf{Future Work:} 
Future work should focus on evaluating KAN's performance in more complex and challenging datasets, such as CIFAR-100, ImageNet, or specialized domains like medical imaging. Additionally, optimizing KAN's computational efficiency and exploring hybrid models that combine KAN with other architectures could further enhance its applicability. Other directions include investigating the role of regularization techniques, such as dropout or weight decay, to better control overfitting in KAN-based models. Furthermore, exploring KAN's performance in transfer learning tasks where rapid convergence is critical could provide additional insights into its utility.

\bibliographystyle{unsrtnat}
\bibliography{bibliography}

\end{document}